%% file: root.tex
\documentclass[letterpaper, 10 pt, conference]{ieeeconf}  

\IEEEoverridecommandlockouts                              

\usepackage{graphicx}
\usepackage{dirtytalk}
\usepackage{etoolbox}
\usepackage{booktabs}
\usepackage{multirow}

\usepackage[mathscr]{euscript}

\usepackage[table,xcdraw]{xcolor}

\usepackage{hyperref}

\usepackage{amsmath}
\usepackage{amssymb}

\usepackage{algorithm}
\usepackage{algpseudocode}
\usepackage{enumerate}

\usepackage{threeparttable}
\usepackage{scalerel}
\usepackage{tikz}
\usepackage{cite}

\title{\LARGE \bf
From Brainwaves to Brain Scans: A Robust Neural Network for EEG-to-fMRI Synthesis
}

\author{Kristofer Roos, Atsushi Fukuda, and Quan Huu Cap
\thanks{All authors are with the AI Development Department, Aillis, Inc., Tokyo, Japan. Corresponding author: Quan Huu Cap ({\tt\small quan.cap@aillis.jp})}
}

\begin{document}

\maketitle
\thispagestyle{empty}
\pagestyle{empty}

\begin{abstract}
    \input{00_Abstract}

\end{abstract}

\begin{keywords}
fMRI reconstruction, functional magnetic resonance imaging, neuroimaging, deep learning, neural networks.
\end{keywords}

\section{INTRODUCTION}
    \input{01_Introduction}

\section{METHODS}
    \input{02_Method}
    
\section{EXPERIMENTAL SETTINGS}
    \input{03_Experimental_Settings}

\section{RESULTS AND DISCUSSION}
    \input{04_Result_Discussion}

\section{CONCLUSION}
    \input{05_Conclusion}


\nocite{*}
\footnotesize{
\bibliographystyle{IEEEtran}
\bibliography{reference}
}

\end{document}

%% file: 00_Abstract.tex
While functional magnetic resonance imaging (fMRI) offers valuable insights into brain activity, it is limited by high operational costs and significant infrastructural demands. 
In contrast, electroencephalography (EEG) provides millisecond-level precision in capturing electrical activity but lacks the spatial fidelity necessary for precise neural localization. 
To bridge these gaps, we propose E2fNet, a simple yet effective deep learning model for synthesizing fMRI images from low-cost EEG data. 
E2fNet is an encoder-decoder network specifically designed to capture and translate meaningful multi-scale features from EEG across electrode channels into accurate fMRI representations. 
Extensive evaluations across three public datasets demonstrate that E2fNet consistently outperforms existing CNN- and transformer-based methods, achieving state-of-the-art results in terms of the structural similarity index measure (SSIM). 
These results demonstrate that E2fNet is a promising, cost-effective solution for enhancing neuroimaging capabilities. 
The code is available at \url{https://github.com/kgr20/E2fNet}. 

%% file: 01_Introduction.tex
\input{figures/tex_files/Fig_1}
Neuroimaging modalities such as functional magnetic resonance imaging (fMRI) are essential for enhancing the diagnosis and treatment of various neurological conditions. 
fMRI data are based on blood oxygen level-dependent (BOLD) signals, which measure changes in cerebral blood flow and oxygenation as an indirect indicator of neuronal activity. 
While fMRI offers rich spatial detail, it requires complex infrastructure with high operational costs, and is often unavailable in many clinical environments \cite{constable2023challenges}. 
In contrast, electroencephalography (EEG) is a relatively low-cost, non-invasive, and portable modality that directly records electrical activity in the brain with millisecond precision \cite{sturzbecher2012simultaneous, debener2012taking}. 
Despite these advantages, EEG lacks the spatial resolution required for precise localization of hemodynamic activity, limiting its diagnostic and research utility. 

The capability to generate fMRI data from EEG signals offers significant potential for cost-effective, real-time functional mapping in various clinical applications. 
These include resource-limited settings, continuous monitoring of neurological disorders, and the generation of synthetic fMRI images for data augmentation \cite{zhuang2019fmri}. 
Previous studies have demonstrated correlations between fMRI data and specific features of EEG signals \cite{laufs2003eeg,laufs2006bold,chang2013eeg}, highlighting EEG's potential to capture aspects of hemodynamic activity reflected in fMRI. 
Recent advancements in cross-modality synthesis using deep neural networks have further positioned EEG-to-fMRI translation as a promising and emerging research direction \cite{liu2019convolutional, calhas2020eeg, calhas2022eeg, lanzino2024nt}. 

Liu et al. \cite{liu2019convolutional} employed a pair of convolutional neural network (CNN) transcoders that directly map EEG signals to fMRI volumes and vice versa, without relying on explicit hemodynamic or leadfield models. 
In simulated datasets, the method achieved a high correlation between predicted and ground-truth data. 
While this pioneering work demonstrated the feasibility of neural transcoding, further advancements are needed to address the inherent complexities of EEG-to-fMRI synthesis, particularly in achieving robust and generalizable performance across diverse contexts. 

Calhas et al. \cite{calhas2020eeg} expanded the exploration of EEG-to-fMRI synthesis using a broader set of deep learning architectures, including autoencoders, generative adversarial networks (GAN) \cite{goodfellow2014generative} variants, and pairwise-learning approaches. 
Despite encountering the challenges of limited simultaneous EEG-fMRI datasets and variable recording conditions, they demonstrated the viability of generating fMRI-like images from EEG signals. 
Notably, their experiments revealed that while autoencoders tend to yield stable (though sometimes oversimplified) outputs, GAN models can capture richer distributional details but suffer from training instability on relatively small datasets. 
Their latest study further improved the cross-modal mapping paradigm by leveraging topographical attention graphs of EEG electrodes alongside Fourier features \cite{calhas2022eeg}. 
The approach explicitly captures non-local relationships between EEG electrodes, addressing the distributed nature of brain activity and enabling the model to better account for the spatial dynamics underlying fMRI signals. 

Lanzino et al. \cite{lanzino2024nt} introduced the Neural Transcoding Vision Transformer (NT-ViT), leveraging an encoder-decoder design optimized for generating fMRI data from EEG signals. 
The NT-ViT model incorporates a domain-matching sub-module to align latent EEG and fMRI representations, significantly enhancing the coherence and accuracy of image generation. 
The results demonstrate that NT-ViT surpasses traditional methods in producing fMRI images with greater clarity and precision, indicating its potential to enhance multimodal neuroimaging analyses. 

While these efforts have demonstrated initial feasibility, the reported studies either show modest performance with significant room for improvement \cite{liu2019convolutional,calhas2020eeg,calhas2022eeg}, require extensive hyperparameter tuning \cite{calhas2022eeg}, or rely on complex model designs \cite{lanzino2024nt}. 
In this work, we propose E2fNet, an accurate and reliable model inspired by the U-Net architecture \cite{ronneberger15unet} for EEG-to-fMRI synthesis. 
The E2fNet is specifically designed to capture and translate meaningful features from EEG across electrode channels into accurate fMRI representations. 
Our E2fNet model not only significantly outperforms previous CNN-based methods \cite{liu2019convolutional,calhas2022eeg}, by a substantial margin, but also offers a simpler design and better structural similarity index measure (SSIM) \cite{wang04ssim} scores compared to the transformer-based approach \cite{lanzino2024nt}. 
Experimental results demonstrate that E2fNet achieves state-of-the-art in terms of SSIM across three publicly available datasets. 
We make our implementation publicly available to encourage reproducibility and further research. 

%% file: figures/tex_files/Fig_1.tex
\begin{figure*}[!t]
\centering
\includegraphics[width=0.95\textwidth]{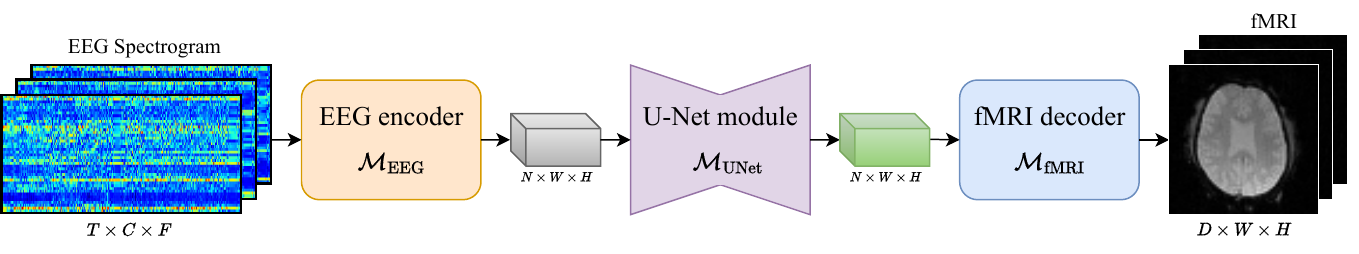}
\caption{
    The overview of our proposed $\mathrm{E2fNet}$ model for EEG to fMRI translation. 
}
\label{fig:fig_1}
\end{figure*}

%% file: 02_Method.tex
Our E2fNet is a fully CNN specifically designed to generate fMRI from EEG data. Let $x \in \mathbb{R}^{T \times C \times F}$ be a spectrogram of multi-channel EEG data, where $T$, $C$, and $F$ correspond to the temporal, electrode channel, and frequency dimension, respectively. 
Let $y \in \mathbb{R}^{D \times W \times H}$ be an fMRI volume representation at a given time, where $D$, $W$, and $H$ correspond to the depth, width, and height. 
Fig. \ref{fig:fig_1} illustrates the overview of the proposed E2fNet. 
Our model consists of an EEG encoder $\mathcal{M}_{\mathrm{EEG}}$, a U-Net module $\mathcal{M}_{\mathrm{UNet}}$, and an fMRI decoder $\mathcal{M}_{\mathrm{fMRI}}$. 

\subsection{The EEG Encoder}
The EEG encoder is built to capture the characteristics of EEG signals effectively. 
Given EEG data \(x \in \mathbb{R}^{T \times C \times F}\), we obtain the encoded features 
\(x_{eeg} = \mathcal{M}_{\mathrm{EEG}}(x)\), where \(x_{eeg} \in \mathbb{R}^{N \times W \times H}\). 
Here, \(N\) denotes the depth (number of feature maps), and \(W \times H\) corresponds to the 
width and height of the target fMRI volume. The design of $\mathcal{M}_{\mathrm{EEG}}$ includes several key principles. 
\textit{First}, $\mathcal{M}_{\mathrm{EEG}}$ progressively expands the temporal dimension from $T \rightarrow N$ (where $N \gg T$), enabling the capture of long-term dependencies and subtle temporal dynamics critical for modeling changes over time. 
\textit{Second}, the model retains the electrode dimension $C$ throughout the convolutional process, preserving the spatial topology and relationships between electrodes. 
This preservation ensures the integrity of spatial information for downstream analyses. 
\textit{Third}, the frequency dimension $F$ is gradually reduced toward the size of $H$. 
This emphasizes the most relevant spectral features while reducing the complexity of the encoded representation. 
All of this can be done by a series of convolutional layers with kernel size of $[1 \times k]$, where $k > 1$, to progressively shrink the $F$ dimension. 

The convolutional process produces a feature tensor of size $N \times C \times H$, which is then resized to $N \times W \times H$ using bicubic interpolation to match the spatial size of the target fMRI. In this work, we designed the $\mathcal{M}_{\mathrm{EEG}}$ to output the feature's depth $N=256$ (i.e., $x_{eeg}\in \mathbb{R}^{256 \times W \times H}$). From our observations, this feature resizing approach is significantly more effective compared to traditional encoders which often compress data into a much lower dimensionality. 

\subsection{The U-Net Module and fMRI Decoder}
The encoded features $x_{eeg} \in \mathbb{R}^{N \times W \times H}$ are then fed into $\mathcal{M}_{\mathrm{UNet}}$ for further processing. 
This U-Net module consists of two down-sample blocks followed by two up-sample blocks, enabling the extraction of multi-scale features. 
Each down-sample block reduces the spatial dimensions by half while the up-sample blocks restore them. 
The number of output channels for each block is $N$ ($N=256$ in this work) and the output of this model is the same as its input, with $x_{unet}=\mathcal{M}_{\mathrm{UNet}}(x_{eeg}) \in \mathbb{R}^{256 \times W \times H}$. 
Here, the extraction of multi-scale features in $\mathcal{M}_{\mathrm{UNet}}$ is crucial, as our preliminary experiments revealed that the model without the U-Net module failed to accurately reconstruct the fMRI targets. 

The fMRI decoder $\mathcal{M}_{\mathrm{fMRI}}$ then gradually reduces the depth of $x_{unet}$ from $N \rightarrow D$ while maintaining the spatial size $W \times H$. 
Finally, the generated fMRI is $\hat{y} = \mathcal{M}_{\mathrm{fMRI}}(x_{unet}) = \mathrm{E2fNet}(x) \in \mathbb{R}^{D \times W \times H}$. 

\subsection{Loss Function}
To train the E2fNet model, we employed the SSIM and mean square error (MSE) losses for balanced structural and pixel-wise accuracy. 
The objective function for training is:
\begin{equation}
\label{eq:1}
\mathcal{L}(y, \hat{y}) = \lambda_{1}(1.0 - \mathrm{SSIM}(y, \hat{y})) + \lambda_{2} \mathrm{MSE}(y, \hat{y}),
\end{equation}
where $y$ and $\hat{y}$ are the ground-truth and generated fMRI, respectively. 
$\lambda_{1}$ and $\lambda_{2}$ are the loss coefficients. 
For more details on the architecture and implementation, please refer to our GitHub repository. 

%% file: 03_Experimental_Settings.tex
\input{tables/Table_I}
\subsection{Data Preprocessing}
We followed the preprocessing method described in the literature \cite{calhas2022eeg}. 
Specifically, given an EEG-fMRI dataset with an EEG sampling rate of $f_s$ (Hz) and an fMRI Time Response (TR) of $f_{TR}$ (second), the EEG signal at each electrode is divided into non-overlapping windows of length $f_s \times f_{TR}$. 
Then, the Fast Fourier Transform (FFT) is applied to extract the frequency components of size $F$ from a 1-dimensional EEG window signal. 
An upper-band filter is applied to the FFT with a cutoff at 250 Hz, and the Direct Current (DC) component is removed. 
As a result, the frequency dimension is $F = 249$. 
For $C$ electrodes in EEG data, the frequency components for a single window have a size of $C \times F$. 
The temporal dimension $T$ of EEG data is set to 20 (i.e., 20 consecutive windows) so that an EEG sample $x \in \mathbb{R}^{20 \times C \times 249}$ corresponds to an fMRI volume $y \in \mathbb{R}^{D \times W \times H}$. 
We normalized fMRI data to range [0, 1] per each volume and the EEG data to range [0, 1] across the entire dataset. For more details on the data pre-processing, please refer to the literature \cite{calhas2022eeg} and our GitHub repository. 

\subsection{Datasets}
In this study, we used three publicly available simultaneous EEG and fMRI datasets namely NODDI \cite{deligianni2014relating}, Oddball \cite{walz2015prestimulus}, and CN-EPFL \cite{pereira2020disentangling}. 
These human subject data were made available in open-access, and no additional ethical approval was required, as confirmed by the license attached to the above open-access data. 
The specific of each dataset is described as follows: 
\subsubsection{NODDI}
A dataset comprising simultaneous EEG-fMRI recordings from 17 healthy adults during resting-state sessions with eyes open. 
EEG was acquired using a 64-channel cap (10-20 system) at a 250Hz sampling rate, capturing millisecond-level neural signals. 
In parallel, fMRI data were collected using a Siemens Avanto 1.5 T clinical scanner with a time response (TR) of 2.16s. 
The resolution of a volume is $30 \times 64 \times 64$. 
Following \cite{lanzino2024nt}, we only processed 15 out of the original 17 participants, resulting in 4,110 paired EEG-fMRI samples. 

\subsubsection{Oddball}
The dataset includes EEG-fMRI recordings from 17 healthy adults performing an oddball paradigm involving auditory and visual tasks. 
EEG was recorded at a 1000 Hz sampling rate across 43 channels, capturing rapid neural oscillations linked to attentional engagement. 
Concurrently, fMRI scans were acquired in a $32 \times 64 \times 64$ volume at a 2-second TR using a Philips Achieva 3T clinical scanner with a single-channel head coil, capturing hemodynamic responses to these rare events. 
This dataset consists of 14,688 paired EEG-fMRI samples. 

\subsubsection{CN-EPFL}
This dataset was collected by the Center for Neuroprosthetics, École Polytechnique Fédérale de Lausanne (EPFL) in Switzerland. 
It comprises concurrent EEG–fMRI recordings from 20 participants, each engaged in a speeded visual discrimination task during data acquisition. 
EEG was collected via a 64-channel cap (10-20 system) at a high sampling rate of 5,000Hz. 
Meanwhile, fMRI scans were obtained at a repetition time (TR) of 1.28 seconds in a $54 \times 108 \times 108$ volume. 
Following \cite{calhas2022eeg}, the discrete cosine transform (DCT II \& III) \cite{ahmed2006discrete} was applied to the original volumes, and subsequently down-sampled to $30 \times 64 \times 64$ for consistency with the other datasets. 
This dataset has a total of 6,880 paired EEG-fMRI samples.

\subsection{Comparison Models}
We compared our E2fNet model with previous methods: CNN-TC \cite{liu2019convolutional}, CNN-TAG \cite{calhas2022eeg}, and NT-ViT \cite{lanzino2024nt}. 
Additionally, we evaluated our model against a GAN-based approach. 
For this, we treated the E2fNet as the generator and employed a binary discriminator to classify an fMRI volume as real or fake. 
We refer to this model as E2fGAN. 
The E2fGAN model was trained using the loss function in Eq. \ref{eq:1} and the GAN loss \cite{goodfellow2014generative}, with the same training settings as in E2fNet. 
Details of the E2fGAN architecture are available in our GitHub repository. 

\subsection{Training and Evaluation}
We trained our E2fNet on the three datasets described above. 
The loss in Eq. \ref{eq:1} with $\lambda_{1} = \lambda_{2} = 0.5$ was employed to train our models. 
The AdamW optimizer \cite{loshchilov2017adamw} was used with the learning rate set to $10^{-3}$. 
The learning rate warm-up was applied for the first 50 training steps. 
All models were trained for 50 epochs on an NVIDIA GeForce RTX 3090 GPU with batch size set to 64. 

In this work, we used the SSIM and peak signal-to-noise ratio (PSNR) metrics to evaluate the quality of the generated results. 
To compare with \cite{lanzino2024nt}, we adopted the leave-one-subject-out cross-validation scheme for the NODDI and Oddball datasets. 
For a dataset with $K$ subjects, we perform $K$ experiments, each time using one subject for evaluation while the remaining subjects are used for training. 
The result is averaged across $K$ experiments. 
For the CN-EPFL dataset, we used the first 16 individuals for training, and the last four for evaluation as in \cite{calhas2022eeg}. 
\input{figures/tex_files/Fig_2}

%% file: tables/Table_I.tex
\begin{table*}[t]
\centering
\caption{Performance comparison of different EEG-to-fMRI generation models on the NODDI, Oddball, and CN-EPFL datasets}
\label{tab:table_1}
\resizebox{0.75\linewidth}{!}{
\begin{tabular}{@{}lllllll@{}}
\toprule
\multirow{2}{*}{Model} & \multicolumn{2}{l}{NODDI}                      & \multicolumn{2}{l}{Oddball}                    & \multicolumn{2}{l}{CN-EPFL}      \\ \cmidrule(l){2-7} 
                       & SSIM                   & PSNR                  & SSIM                   & PSNR                  & SSIM           & PSNR            \\ \midrule
CNN-TC \cite{liu2019convolutional}                 & 0.449 ± 0.060          & ---                   & 0.189 ± 0.038          & ---                   & 0.519          & ---             \\
CNN-TAG \cite{calhas2022eeg}                & 0.472 ± 0.010          & ---                   & 0.200 ± 0.017          & ---                   & 0.522          & ---             \\
NT-ViT \cite{lanzino2024nt}                 & 0.581 ± 0.048          & \textbf{21.56 ± 1.06} & 0.627 ± 0.051          & \textbf{23.33 ± 1.04} & N/A            & ---             \\ \midrule
E2fGAN (\textbf{proposed})                 & 0.576 ± 0.047          & 18.535 ± 1.345        & 0.583 ± 0.034          & 18.711 ± 1.754        & 0.607          & 18.172          \\
E2fNet (\textbf{proposed})                 & \textbf{0.605 ± 0.046} & 20.096 ± 1.280        & \textbf{0.631 ± 0.042} & 22.193 ± 1.013        & \textbf{0.674} & \textbf{22.781} \\ \bottomrule
\end{tabular}
}
\end{table*}

%% file: figures/tex_files/Fig_2.tex
\begin{figure*}[!t]
\centering
\includegraphics[width=0.99\textwidth]{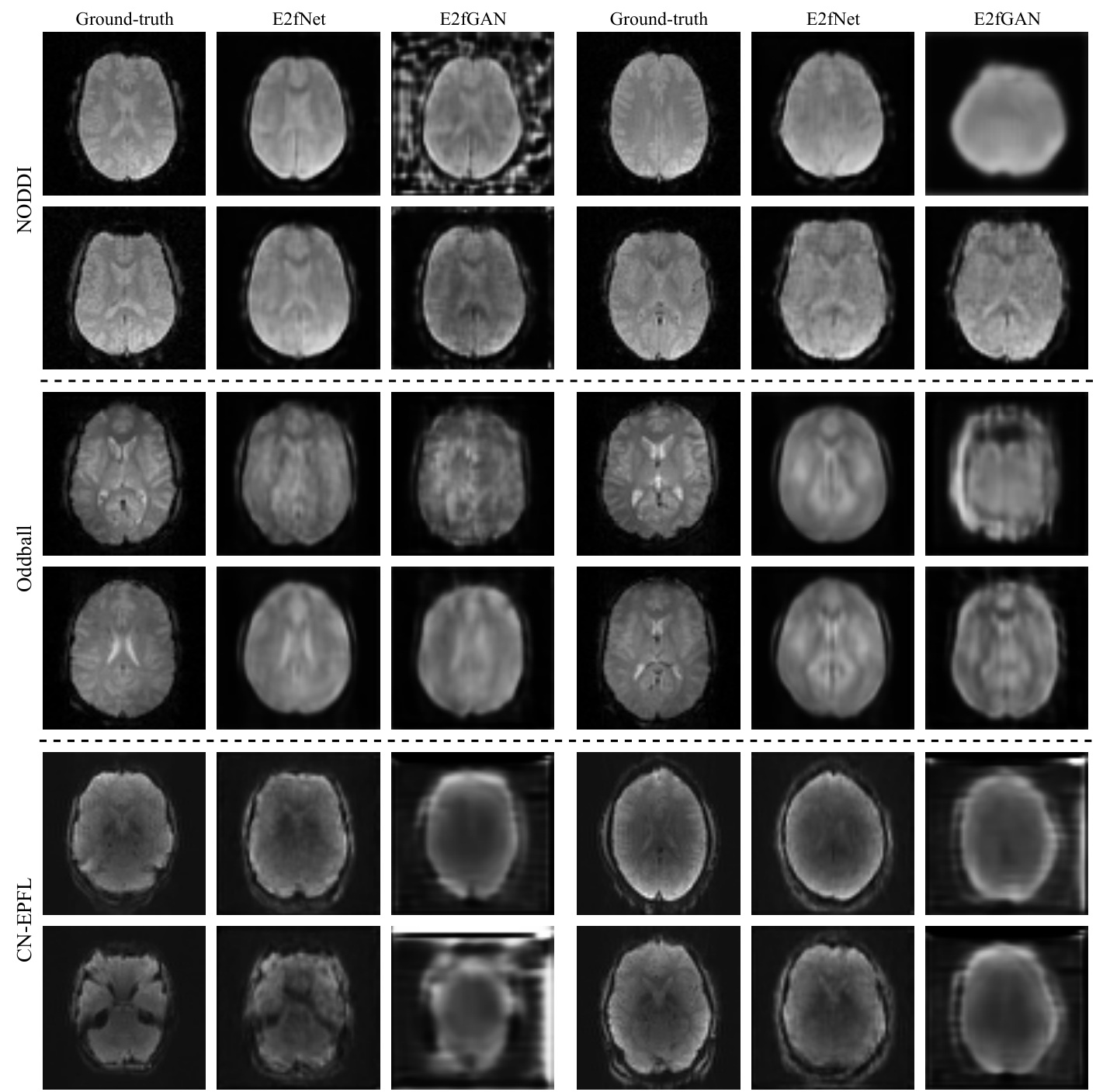}
\caption{
    Visual comparison of the generated fMRI by E2fNet and E2fGAN models on the NODDI, Oddball, and CN-EPFL datasets. 
}
\label{fig:fig_2}
\end{figure*}

%% file: 04_Result_Discussion.tex
Table \ref{tab:table_1} presents the performance comparison of different models on the three datasets. 
Our proposed E2fNet achieved state-of-the-art results in terms of SSIM across all benchmarks, with SSIM scores of 0.605 on NODDI, 0.631 on Oddball, and 0.674 on CN-EPFL datasets. 
The second-best performing model was NT-ViT, followed by our E2fGAN, with the CNN-TAG and CNN-TC models ranking lower.

In comparison to the previous CNN-TC and CNN-TAG models, our models significantly improved the fMRI reconstruction performance by substantial margins. 
Most notably on the Oddball dataset with a 0.442 SSIM increment over the CNN-TC \cite{liu2019convolutional} model (from 0.189 to 0.631), marking a considerable leap in performance. 
The CNN-TAG and CNN-TC models, in contrast, were less effective in capturing meaningful features from EEG data for generating fMRI targets. 

The strong performance of E2fNet can be attributed to two primary innovations: the design of the EEG encoder $\mathcal{M}_{\mathrm{EEG}}$ and the integration of the U-Net module $\mathcal{M}_{\mathrm{UNet}}$. 
The encoder effectively captures meaningful features across EEG electrodes, while the U-Net module enables the extraction of multi-scale features. 
We observed that the model without the U-Net module failed to accurately reconstruct the fMRI targets, with relatively low SSIM scores (e.g., below 0.45 on the NODDI dataset). 

Compared to the NT-ViT model, E2fNet achieved higher SSIM scores but lower PSNR on the NODDI and Oddball datasets (see Table \ref{tab:table_1}). 
The NT-ViT’s strong PSNR results may be attributed to the use of the Vision Transformer architecture \cite{dosovitskiy21vit}. 
However, we argue that SSIM is a more relevant metric for assessing perceptual quality from a human visual perspective, which is critical in medical imaging applications. 
Furthermore, E2fNet offers a much simpler design compared to NT-ViT. 
Specifically, our E2fNet follows a simple encoder-decoder architecture design whereas the NT-ViT employed two transformer-based encoder-decoder networks. 

To provide a more comprehensive evaluation, we compared E2fNet with the GAN-based E2fGAN model, which uses E2fNet as the generator. 
Fig. \ref{fig:fig_2} presents a visual comparison of E2fNet and E2fGAN across the three datasets. 
During the experiments, we observed that training the GAN model was often unstable, leading to unexpected outputs (Fig. \ref{fig:fig_2}, first row of the NODDI and Oddball datasets, first example in the second row of the CN-EPFL dataset). 
This instability contributed to E2fGAN's lower SSIM and PSNR scores compared to E2fNet. 
Nevertheless, in some cases, E2fGAN produced visually sharper fMRI volumes. 
Future work could explore improvements by adopting more robust GAN training strategies. 
It is important to note that SSIM and PSNR metrics may sometimes be less reliable when evaluating small or blurry images. 
For instance, while E2fGAN frequently generated oversmoothed and less detailed results on the CN-EPFL dataset, it still achieved seemingly reasonable SSIM and PSNR scores. 
We believe developing better similarity metrics could help address this limitation. 

Another potential direction is to further develop a model that not only works with scalp EEG but also with subcortical signals. 
We believe this is possible to learn such patterns from data, as prior work has shown that subcortical signals can be indirectly detected through EEG \cite{seeber2019subcortical}. 
There is still room for improvement, and we plan to investigate these aspects in future work. 

%% file: 05_Conclusion.tex
In this study, we proposed E2fNet, a simple yet effective model for EEG-to-fMRI generation. 
E2fNet effectively captures meaningful EEG features and consistently demonstrates strong performance across multiple datasets. 
While further studies are necessary to fully validate the practical utility of the generated fMRI data, we believe that our approach holds promising potential for cost-effective applications in the field of neuroimaging. 

%% file: root.bbl
\begin{thebibliography}{10}
\providecommand{\url}[1]{#1}
\csname url@samestyle\endcsname
\providecommand{\newblock}{\relax}
\providecommand{\bibinfo}[2]{#2}
\providecommand{\BIBentrySTDinterwordspacing}{\spaceskip=0pt\relax}
\providecommand{\BIBentryALTinterwordstretchfactor}{4}
\providecommand{\BIBentryALTinterwordspacing}{\spaceskip=\fontdimen2\font plus
\BIBentryALTinterwordstretchfactor\fontdimen3\font minus \fontdimen4\font\relax}
\providecommand{\BIBforeignlanguage}[2]{{%
\expandafter\ifx\csname l@#1\endcsname\relax
\typeout{** WARNING: IEEEtran.bst: No hyphenation pattern has been}%
\typeout{** loaded for the language `#1'. Using the pattern for}%
\typeout{** the default language instead.}%
\else
\language=\csname l@#1\endcsname
\fi
#2}}
\providecommand{\BIBdecl}{\relax}
\BIBdecl

\bibitem{constable2023challenges}
R.~T. Constable, ``Challenges in fmri and its limitations bt—functional neuroradiology: Principles and clinical applications,'' \emph{Functional Neuroradiology}, pp. 497--510, 2023.

\bibitem{sturzbecher2012simultaneous}
M.~J. Sturzbecher and D.~B. de~Araujo, ``Simultaneous eeg-fmri: integrating spatial and temporal resolution,'' in \emph{The Relevance of The Time Domain to Neural Network Models}, 2012, pp. 199--217.

\bibitem{debener2012taking}
S.~Debener, F.~Minow, R.~Emkes, K.~Gandras, and M.~De~Vos, ``How about taking a low-cost, small, and wireless eeg for a walk?'' \emph{Psychophysiology}, vol.~49, no.~11, pp. 1617--1621, 2012.

\bibitem{zhuang2019fmri}
P.~Zhuang, A.~G. Schwing, and O.~Koyejo, ``Fmri data augmentation via synthesis,'' in \emph{IEEE International Symposium on Biomedical Imaging}, 2019, pp. 1783--1787.

\bibitem{laufs2003eeg}
H.~Laufs, A.~Kleinschmidt, A.~Beyerle, E.~Eger, A.~Salek-Haddadi, C.~Preibisch, and K.~Krakow, ``Eeg-correlated fmri of human alpha activity,'' \emph{Neuroimage}, vol.~19, no.~4, pp. 1463--1476, 2003.

\bibitem{laufs2006bold}
H.~Laufs, J.~L. Holt, R.~Elfont, M.~Krams, J.~S. Paul, K.~Krakow, and A.~Kleinschmidt, ``Where the bold signal goes when alpha eeg leaves,'' \emph{Neuroimage}, vol.~31, no.~4, pp. 1408--1418, 2006.

\bibitem{chang2013eeg}
C.~Chang, Z.~Liu, M.~C. Chen, X.~Liu, and J.~H. Duyn, ``Eeg correlates of time-varying bold functional connectivity,'' \emph{Neuroimage}, vol.~72, pp. 227--236, 2013.

\bibitem{liu2019convolutional}
X.~Liu and P.~Sajda, ``A convolutional neural network for transcoding simultaneously acquired eeg-fmri data,'' in \emph{International IEEE/EMBS Conference on Neural Engineering}, 2019, pp. 477--482.

\bibitem{calhas2020eeg}
D.~Calhas and R.~Henriques, ``Eeg to fmri synthesis: Is deep learning a candidate?'' in \emph{International Conference on Information Systems Development}, 2023, pp. 1--12.

\bibitem{calhas2022eeg}
------, ``Eeg to fmri synthesis benefits from attentional graphs of electrode relationships,'' in \emph{Machine Learning for Healthcare Conference}, vol. 219, 2023, pp. 1--23.

\bibitem{lanzino2024nt}
\BIBentryALTinterwordspacing
R.~Lanzino, F.~Fontana, L.~Cinque, F.~Scarcello, and A.~Maki, ``Nt-vit: Neural transcoding vision transformers for eeg-to-fmri synthesis,'' \emph{arXiv:2409.11836}, 2024. [Online]. Available: \url{https://arxiv.org/abs/2409.11836}
\BIBentrySTDinterwordspacing

\bibitem{goodfellow2014generative}
I.~Goodfellow, J.~Pouget-Abadie, M.~Mirza, B.~Xu, D.~Warde-Farley, S.~Ozair, A.~Courville, and Y.~Bengio, ``Generative adversarial nets,'' in \emph{Advances in Neural Information Processing Systems}, 2014, pp. 2672--2680.

\bibitem{ronneberger15unet}
O.~Ronneberger, P.~Fischer, and T.~Brox, ``U-net: Convolutional networks for biomedical image segmentation,'' in \emph{International Conference on Medical Image Computing and Computer-Assisted Intervention}, 2015, pp. 234--241.

\bibitem{wang04ssim}
Z.~Wang, A.~C. Bovik, H.~R. Sheikh, and E.~P. Simoncelli, ``Image quality assessment: from error visibility to structural similarity,'' \emph{IEEE Transactions on Image Processing}, vol.~13, no.~4, pp. 600--612, 2004.

\bibitem{deligianni2014relating}
F.~Deligianni, M.~Centeno, D.~W. Carmichael, and J.~D. Clayden, ``Relating resting-state fmri and eeg whole-brain connectomes across frequency bands,'' \emph{Frontiers in Neuroscience}, vol.~8, p. 258, 2014.

\bibitem{walz2015prestimulus}
J.~M. Walz, R.~I. Goldman, M.~Carapezza, J.~Muraskin, T.~R. Brown, and P.~Sajda, ``Prestimulus eeg alpha oscillations modulate task-related fmri bold responses to auditory stimuli,'' \emph{NeuroImage}, vol. 113, pp. 153--163, 2015.

\bibitem{pereira2020disentangling}
M.~Pereira, N.~Faivre, I.~Iturrate, M.~Wirthlin, L.~Serafini, S.~Martin, A.~Desvachez, O.~Blanke, D.~Van De~Ville, and J.~d.~R. Mill{\'a}n, ``Disentangling the origins of confidence in speeded perceptual judgments through multimodal imaging,'' \emph{Proceedings of the National Academy of Sciences}, vol. 117, no.~15, pp. 8382--8390, 2020.

\bibitem{ahmed2006discrete}
N.~Ahmed, T.~Natarajan, and K.~R. Rao, ``Discrete cosine transform,'' \emph{IEEE Transactions on Computers}, vol. 100, no.~1, pp. 90--93, 2006.

\bibitem{loshchilov2017adamw}
I.~Loshchilov and F.~Huntter, ``Decoupled weight decay regularization,'' in \emph{International Conference on Learning Representations}, 2019, pp. 1--8.

\bibitem{dosovitskiy21vit}
A.~Dosovitskiy, L.~Beyer, A.~Kolesnikov, D.~Weissenborn, X.~Zhai, T.~Unterthiner, M.~Dehghani, M.~Minderer, G.~Heigold, S.~Gelly, J.~Uszkoreit, and N.~Houlsby, ``An image is worth 16x16 words: Transformers for image recognition at scale,'' in \emph{International Conference on Learning Representations}, 2021, pp. 1--22.

\bibitem{seeber2019subcortical}
M.~Seeber, L.-M. Cantonas, M.~Hoevels, T.~Sesia, V.~Visser-Vandewalle, and C.~M. Michel, ``Subcortical electrophysiological activity is detectable with high-density eeg source imaging,'' \emph{Nature Communications}, vol.~10, no.~1, p. 753, 2019.

\end{thebibliography}
